\documentclass[11pt]{article}

\usepackage[utf8]{inputenc}
\usepackage[T1]{fontenc}
\usepackage{bm}

\usepackage[margin=1in]{geometry}
\usepackage{parskip}

\usepackage{amsmath,amssymb}

\usepackage{graphicx}
\usepackage{xcolor}
\definecolor{RAGMuted}{RGB}{90,90,90}
\definecolor{RAGBorder}{RGB}{200,200,200}
\definecolor{RAGUrlBlue}{RGB}{0,0,128}

\usepackage{booktabs}
\usepackage{multirow}
\usepackage{array}
\usepackage{tabularx}
\usepackage{caption}

\newlength{\TableWidth}
\usepackage{placeins}

\usepackage[most]{tcolorbox}
\usepackage{fvextra}
\usepackage{ragged2e}
\newcommand{\maybeicon}[1]{\IfFileExists{#1}{\includegraphics[height=2.2ex]{#1}}{}}

\usepackage[
  colorlinks=true,
  linkcolor=blue,
  citecolor=blue,
  urlcolor=RAGUrlBlue
]{hyperref}
\usepackage{xurl}
\newcommand{\sig}[1]{\ifmmode\mathbf{#1}\else\textbf{#1}\fi}

\newcommand{\ChiSize}{\normalsize}
\newcommand{\RowLabelSize}{\normalsize}
\newcommand{\PvalSize}{\footnotesize}

\newcommand{\rowlabel}[1]{{\bfseries\RowLabelSize #1}}

\newcommand{\statcellNS}[2]{%
  \begin{tabular}[c]{@{}c@{}}%
    {\ChiSize #1}\\[-1pt]%
    {\PvalSize\textcolor{black}{#2}}%
  \end{tabular}%
}

\newcommand{\statcellSig}[2]{%
  \begin{tabular}[c]{@{}c@{}}%
    {\ChiSize\boldmath\bfseries #1}\\[-1pt]%
    {\PvalSize\boldmath\bfseries\textcolor{black}{#2}}%
  \end{tabular}%
}

\title{Legal RAG Bench: an end-to-end benchmark for legal RAG}
\author{
    Abdur-Rahman Butler, Umar Butler \\
    Isaacus
}
\date{2 March 2026}

\begin{document}
\maketitle

\begin{abstract}
We introduce Legal RAG Bench, a benchmark and evaluation methodology for assessing the end-to-end performance of legal RAG systems. As a benchmark, Legal RAG Bench consists of 4,876 passages from the Victorian Criminal Charge Book alongside 100 complex, hand-crafted questions demanding expert knowledge of criminal law and procedure. Both long-form answers and supporting passages are provided. As an evaluation methodology, Legal RAG Bench leverages a full factorial design and novel hierarchical error decomposition framework, enabling apples-to-apples comparisons of the contributions of retrieval and reasoning models in RAG. We evaluate three state-of-the-art embedding models (Isaacus' Kanon 2 Embedder, Google's Gemini Embedding 001, and OpenAI's Text Embedding 3 Large) and two frontier LLMs (Gemini 3.1 Pro and GPT-5.2), finding that information retrieval is the primary driver of legal RAG performance, with LLMs exerting a more moderate effect on correctness and groundedness. Kanon 2 Embedder, in particular, had the largest positive impact on performance, improving average correctness by 17.5 points, groundedness by 4.5 points, and retrieval accuracy by 34 points. We observe that many errors attributed to hallucinations in legal RAG systems are in fact triggered by retrieval failures, concluding that retrieval sets the ceiling for the performance of many modern legal RAG systems. We document why and how we built Legal RAG Bench alongside the results of our evaluations. We also openly release our code and data to assist with reproduction of our findings.
\end{abstract}

\section{Introduction}
Retrieval-augmented generation (RAG) involves the conditioning of LLMs on current context retrieved from corpora. In recent years, RAG has become one of the most popular approaches for injecting new information into AI systems, including especially legal AI systems, which often demand elevated levels of groundedness and verifiability. Despite the strong uptake of RAG by the legal tech industry, however, high-quality end-to-end benchmarks for legal RAG applications remain relatively scarce. Most public benchmarks either target general applications or assess only one component of legal RAG, often one of retrieval and text generation. More broadly, many legal AI benchmarks suffer from poor label quality, methodological and design flaws, and general mismatches between what they are purported to evaluate and what they actually evaluate.

This lack of high-quality end-to-end legal RAG evaluations means that application builders are often forced to rely on assessments that only weakly correlate, if at all, with real-world performance, in turn reducing the value of their applications to lawyers and end consumers. Correspondingly, low-quality legal benchmarks can lead to model builders sacrificing real-world performance for an unworkable model of what ``good legal AI'' should look like.

Thus, to address the shortcomings of most legal RAG benchmarks and thereby elevate the overall quality of legal applications built on legal benchmarks, we are introducing Legal RAG Bench, a new dataset and evaluation methodology for assessing the end-to-end performance of legal RAG systems. As a dataset, Legal RAG Bench consists of 4,876 passages from the Victorian Criminal Charge Book \cite{jcv_criminal_charge_book} alongside 100 expert-crafted, meaningfully challenging questions demanding expert-level knowledge of Victorian criminal law and procedure. Each question is paired with a long-form answer and a supporting passage, enabling joint assessments of both retrieval and answer generation performance. As a methodology, Legal RAG Bench runs a full factorial experiment over multiple frontier retrieval and generative models, leveraging a novel error taxonomy to decompose errors into hallucinations, retrieval failures, and reasoning failures and then attribute those errors to particular models. Our analysis finds conclusively that retrieval quality is the primary driver of end-to-end legal RAG performance and that most hallucinations in production legal RAG systems are induced by retrieval failures. We openly release the data, code, and results behind Legal RAG Bench to support reproducible evaluation.

\section{Related work}
Legal AI evaluations remain scarce relative to other domains. Of the evaluations that do exist, many suffer from poor label quality, methodological flaws, and flawed assumptions about their real-world value. Underlying these issues is often a more serious failure to properly involve and rely on legal subject-matter expertise in the design and annotation of benchmarks.

\subsection{Retrieval benchmarks}
The legal retrieval evaluation space has not changed much in recent times, barring the release of the Massive Legal Embedding Benchmark (MLEB) \cite{butler2025mleb} in October 2025. To date, MLEB remains the largest, most diverse, and most comprehensive open-source legal benchmark for retrieval models. In introducing MLEB, we contrasted how MLEB was constructed with other popular retrieval benchmarks, highlighting serious failings across the board, including in the legal split of the Massive Text Embedding Benchmark (MTEB) \cite{muennighoff2023mtebmassivetextembedding}.

Consider, for example, the AILA Casedocs and AILA Statutes \cite{bhattacharya-2019-overview} datasets included in MTEB. We found that they had been created using an ``automated methodology'' that paired ``facts stated in certain [Indian] Supreme Court cases'' with cases and statutes that had been ``cited by the lawyers arguing those cases'' \cite{bhattacharya-2019-overview}. According to the authors, ``actually involving legal experts (e.g., to find relevant prior cases / statutes) would have required a significant amount of financial resources and time'' \cite[p.~4]{bhattacharya-2019-overview}. Given a basic understanding of how judgments are written and how legal citations work, it is clear that retrieval of the facts of a case based solely on the text of a judgment that cited that case is impossible in most cases and, even where possible, is of little to no practical value. By way of illustration, although the case of \textit{Donoghue v Stevenson} \cite{donoghue1932} was factually about a certain May Donoghue who had fallen ill from drinking a ginger beer that contained a decomposed snail, that case has been cited and continues to be cited in judgments all around the world in support of points of law that, objectively, have absolutely zero relevance to snails, ginger beer, or Ms Donoghue.

\subsection{LLM and RAG benchmarks}
Similar flaws are observable even in the most popular text generation and RAG benchmarks, including, for example, Humanity's Last Exam (HLE) \cite{hle-2026}, an evaluation set said to be ``the final closed-ended academic benchmark of its kind with broad subject coverage'', having been ``developed globally by subject-matter experts'' at a cost of at least \$500,000. Our review of HLE's legal subset revealed that most examples were either inappropriate, poorly framed, or mislabeled. The question-answer pair shown below is a prime example \cite{cais_hle_dataset}.

\begin{tcolorbox}[title={Humanity's Last Exam example \#6736f7bc980211368f0f94eb}, colback=white, colframe=black!30, sharp corners, breakable]
\textbf{Question:}
\begin{Verbatim}[
  breaklines=true,
  breakanywhere=true,
  breaksymbolleft={},
  breaksymbolright={}
]
Tommy brought a plot of land after retirement. Tommy split the land with his brother James in 2016.

The lawyer filed and recorded 2 Deeds. There's two trailers on the land. There's two different Tax Map Numbers for the land and trailers.

The land now has Lot A filed and recorded legally owned by Tommy and Lot B filed and recorded legally owned by James.
Tommy number is 1234567890 and trailer 1234567890.1. Lot A

James number is 0987654321 and 0987654321.1. Lot B

Tommy died in 2017. Tommy's Estate had to correct Tommy' Deed as Lot A was incorrectly filed. Tommy Deed was filed and re-recorded to Lot B in 2017.

Now Tommy number is 1234567890 and trailer 1234567890.1 Lot B.

James took a lien out with his Deed in 2016, for Lot B.

James never corrected his Deed to Lot A, as advised by his lawyer in 2017. James needed his original Deed Lot B for correction.

Neither Tommy or James trailers never moved.

Who owns Lot A and Lot B with what number?
\end{Verbatim}

\medskip
\textbf{Answer:}
\begin{Verbatim}[
  breaklines=true,
  breakanywhere=true,
  breaksymbolleft={},
  breaksymbolright={}
]
Tommy owns Lot A and Lot B with 1234567890.
\end{Verbatim}
\end{tcolorbox}

Not only is this question flawed, but the purported answer is also incorrect. For one, the question does not provide sufficient context for it to be wholly answerable. Different jurisdictions have different rules on how ownership of property is transferred and recognized. The question, however, does not state the applicable jurisdiction nor does it provide rules under which it is to be interpreted. Secondly, the purported answer is, regardless, almost certainly incorrect. By virtue of Tommy's estate having corrected his deed to record Lot B and not Lot A as corresponding to his plot of land, Tommy cannot simultaneously own both lots.

Failures to make the assumptions upon which an answer depends reasonably inferable from all the context provided to a model are rife in open-source legal evaluation datasets. But even where there are no obvious labeling or methodological errors, there can still be fatal mismatches between what a benchmark is portrayed as evaluating and what it effectively actually evaluates. Both LegalBench \cite{guha2023legalbenchcollaborativelybuiltbenchmark} and LegalBench-RAG \cite{pipitone2024legalbenchragbenchmarkretrievalaugmentedgeneration} suffer from the latter problem. Despite being marketed as valuable stress tests of the reasoning and retrieval capabilities of LLMs, the vast majority of their data is in fact comprised of low-value, relatively trivial text classification and sentiment analysis tasks requiring simple yes or no answers, examples of which include questions such as, ``Does the clause describe a license grant to a licensee (incl. sublicensor) and the affiliates of such licensee/sublicensor?'' and ``Consider the Cologuard Promotion Agreement between Exact Sciences Corporation and Pfizer Inc.; Does this contract include any right of first refusal, right of first offer, or right of first negotiation?''.

Similarly, Stanford RegLab's HousingQA and BarExamQA \cite{Zheng_2025} evaluations, although far more meaningfully challenging than LegalBench \cite{guha2023legalbenchcollaborativelybuiltbenchmark} and LegalBench-RAG \cite{pipitone2024legalbenchragbenchmarkretrievalaugmentedgeneration}, still fail to actually evaluate the capabilities they claim to. In particular, by evaluating models exclusively on closed-ended questions, they are incapable of simulating the far more chaotic conditions of real-world legal RAG systems where LLMs can not only fail to produce coherent answers but can also sometimes hallucinate correct yet fundamentally ungrounded answers. The authors themselves acknowledge this shortcoming when they state, ``Our datasets are restricted in subject-matter domains and restricted to multiple-choice answer forms to enable automatic evaluation of the downstream task. They may not represent a `realistic' approximation of the full natural distribution of legal questions'' \cite{Zheng_2025}.

\section{Dataset}
To address the full spectrum of problems plaguing contemporary legal RAG evaluations, we are introducing Legal RAG Bench, a benchmark and evaluation methodology for the robust evaluation of the real-world end-to-end performance of legal RAG systems.

As a dataset, Legal RAG Bench consists of 4,876 passages sampled from the Judicial College of Victoria’s Criminal Charge Book \cite{jcv_criminal_charge_book} paired with 100 complex, meaningfully challenging, hand-crafted questions demanding expert-level knowledge of Victorian criminal law and procedure to be answered correctly. In this respect, Legal RAG Bench represents the first evaluation set to assess the performance of retrieval and generative models in larger RAG systems aimed at providing practical, real-world legal advice, particularly in an under-resourced but vitally important domain, namely, criminal law.

Uniquely, subject-matter expertise in law and AI informed and guided every stage of Legal RAG Bench’s design and development, from the crafting of a diverse set of realistic hypothetical scenarios all the way to the selection of Victoria’s Criminal Charge Book as the foundation of the benchmark given its central relevance to the day-to-day work of criminal lawyers.

In constructing Legal RAG Bench, we downloaded each section of the Criminal Charge Book \cite{jcv_criminal_charge_book} as Microsoft Word documents and converted them into Markdown. We leveraged a complex set of heuristics to break sections up into their full hierarchy, such as chapters and subchapters, and then, where necessary, we further chunked sections using the semchunk \cite{isaacus_semchunk} semantic chunking algorithm such that no chunk was over 512 tokens in length as determined by the Kanon legal tokenizer.

After building our corpus of 4,876 passages, we randomly sampled passages, hand-crafting 100 complex, meaningfully challenging questions that, to the maximum extent possible, would require each of those passages alone to be answered correctly. In drafting questions, we made them as lexically dissimilar from relevant passages as possible in order to stress test the semantic understanding of evaluated models.

Long-form answers were also hand-crafted for each question by subject-matter experts, yielding question-answer-evidence triplets for each example in Legal RAG Bench. This unique structure allows for the retrieval and generative components of legal RAG systems to be evaluated both in isolation and jointly.

\section{Evaluation}
As a methodology, Legal RAG Bench constitutes the first full factorial experiment evaluating the retrieval accuracy, correctness, and groundedness of legal RAG systems, enabling empirical apples-to-apples assessments of the relative impact of retrieval and generative models on performance. Specifically, we evaluate every combination of three state-of-the-art embedding models, Isaacus' Kanon 2 Embedder, Google's Gemini Embedding 001, and OpenAI's Text Embedding 3 Large, and two frontier LLMs, Gemini 3.1 Pro and GPT-5.2. These models were selected based on their popularity and purported performance at legal retrieval and reasoning. To minimize confounding variables, we use the same barebones Langchain-based \cite{chase2026langchain} RAG pipeline across all models without modifying any hyperparameters from their defaults. We leverage GPT-5.2 in high reasoning mode as a ``judge'' to grade the performance of LLMs against answers and retrieved contexts. Our internal review of our LLM-as-a-Judge evaluation pipeline found that using GPT-5.2 in high reasoning mode as a judge resulted in 99\% accuracy, which we assessed to be tolerable. We were able to achieve such high accuracy by providing our judge with a clear evaluation rubric and a simple set of binary outcomes to predict that were as unambiguous as possible.

In particular, for each question $i$, embedding model $e$, and LLM $l$, we assessed the following evaluation dimensions:
\begin{enumerate}
  \item \textbf{Correctness} ($c_{eli}$): $1$ if the model's answer entails the reference answer; $0$ otherwise.
  \item \textbf{Groundedness} ($g_{eli}$): $1$ if the answer is supported by the retrieved passages provided to the model (irrespective of whether those passages are actually relevant); $0$ otherwise.
  \item \textbf{Retrieval accuracy} ($r_{ei}$): $1$ if the annotated supporting passage is retrieved by the embedding model; $0$ otherwise.
\end{enumerate}

Together, these three signals give visibility into the effectiveness of each component of a RAG system, making it possible to detect failure modes that would be obscured by looking only at overall RAG accuracy, such as hallucinations. Indeed, any serious evaluation of RAG performance must account for hallucinations, including hallucinations that nevertheless produce correct answers. Legal research and analysis is very much evidence-driven—the verifiability of legal conclusions is often just as important as the veracity of those conclusions, if not more so. Untrue but verifiable conclusions can always be proved to be false, yet true but unverifiable conclusions can never be proven to be true.

Thus, to account for failure modes such as hallucinations that happen to yield correct answers, we further introduce a new hierarchical error decomposition taxonomy consisting of the following error types:
\begin{enumerate}
    \item \textbf{Hallucination}: where a generative model invents facts not in its provided context ($g_{eli} = 0$).
    \item \textbf{Retrieval error}: where an embedding model fails to retrieve a relevant passage, yielding a grounded but incorrect answer from the generative model ($g_{eli} = 1 \wedge c_{eli} = 0 \wedge r_{ei} = 0$).
    \item \textbf{Reasoning error}: where an embedding model retrieves a relevant passage, but the generative model nevertheless generates an incorrect answer ($g_{eli} = 1 \wedge c_{eli} = 0 \wedge r_{ei} = 1$).
\end{enumerate}

\begin{figure}[!htbp]
  \centering
  \includegraphics[width=0.75\linewidth]{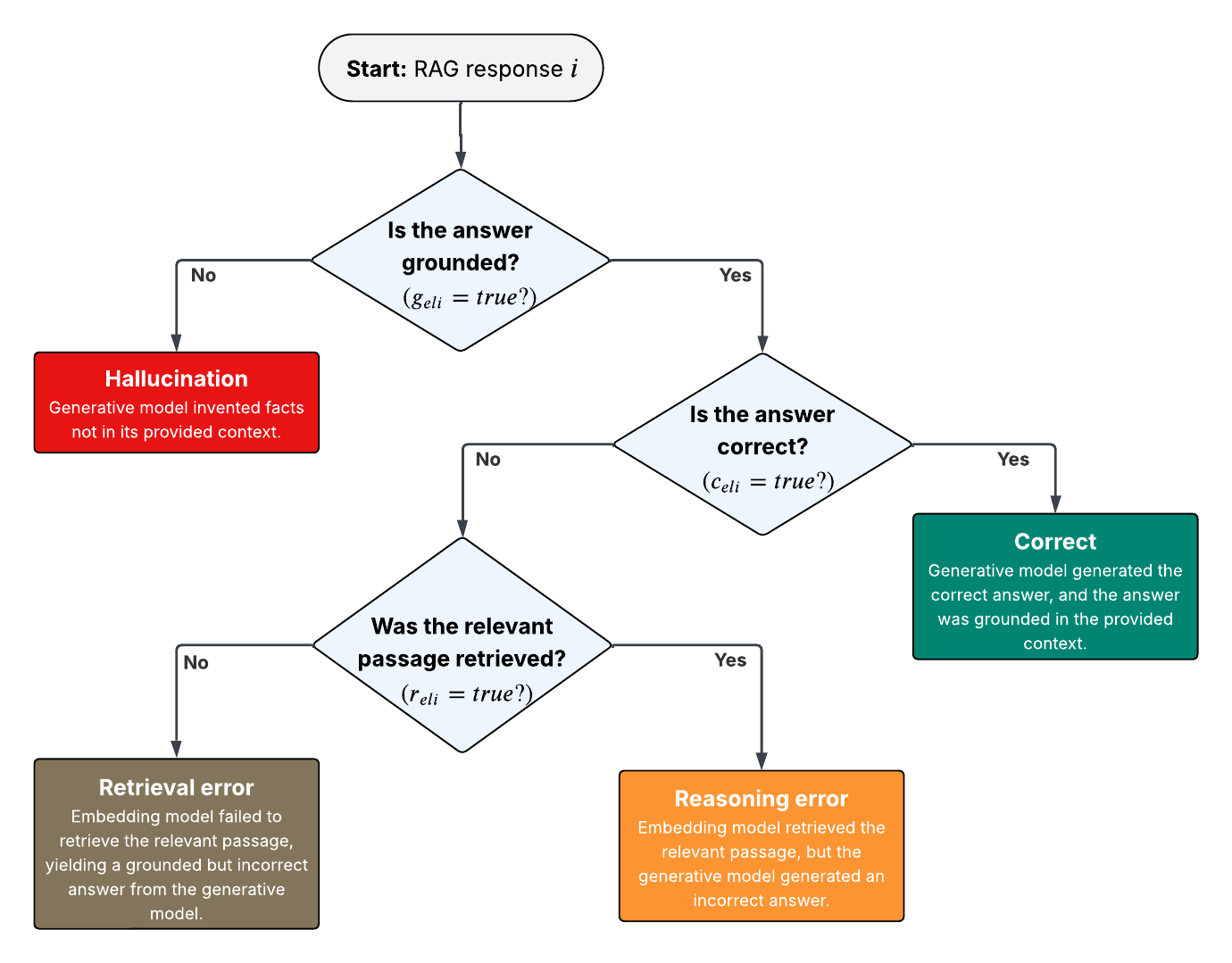}
  \caption{A flow chart representing our hierarchical RAG error decomposition taxonomy.}
  \label{fig:error-taxonomy}
\end{figure}
In our taxonomy, hallucinations are deliberately treated as the first possible failure mode in a RAG pipeline, ahead even of retrieval failures, because, as previously mentioned, in the real world, hallucinated or ungrounded answers make it impossible to verify the correctness of those answers on their own. We also only recognize retrieval failures that have yielded incorrect answers because, logically, if a generative model's response is grounded exclusively in its retrieved context and that response is a correct answer, then it follows that the retrieved context is likely to be relevant, even if it does not happen to include the specific passage Legal RAG Bench labels as most relevant given that it is possible for other passages to also be relevant.

\section{Results}
\subsection{Evaluation dimensions}
\autoref{tbl:dims-by-llm-emb} reports the mean correctness, groundedness, and retrieval accuracy of each evaluated combination of embedding model and LLM. \autoref{tbl:avg-emb-llm-dims} reports corresponding model-level averages where the scores of embedding models have been averaged across LLMs and vice versa for LLMs.

\begin{table}[t]
\centering
\small
\renewcommand{\arraystretch}{1.15}

\begin{tabularx}{\TableWidth}{@{}
  >{\raggedright\arraybackslash}p{\dimexpr0.24\TableWidth\relax}
  >{\raggedright\arraybackslash}X
  c c c
@{}}
\toprule
LLM & Embedder & Correct. (\%) & Grounded. (\%) & Retrieval acc. (\%) \\
\midrule

\multirow{3}{=}{Gemini 3.1 Pro}
  & Kanon 2 Emb.        & \textbf{95.0} & 95.0          & \textbf{86.0} \\
  & Text Emb. 3 L.      & 77.0          & \textbf{96.0} & 52.0          \\
  & Gem. Emb. 001     & 75.0          & 92.0          & 53.0          \\
\midrule

\multirow{3}{=}{GPT-5.2}
  & Kanon 2 Emb.        & \textbf{93.0} & \textbf{97.0} & \textbf{86.0} \\
  & Text Emb. 3 L.      & 76.0          & 87.0          & 52.0          \\
  & Gem. Emb. 001     & 73.0          & 82.0          & 53.0          \\
\bottomrule
\end{tabularx}
\captionsetup{width=\TableWidth, skip=6pt, justification=centering}
\caption{Dimension scores by LLM by embedding model.}
\label{tbl:dims-by-llm-emb}
\end{table}

\FloatBarrier

\begin{table}[!htbp]
\centering
\small
\renewcommand{\arraystretch}{1.15}

\begin{tabularx}{\TableWidth}{@{}
  >{\raggedright\arraybackslash}X
  c c c
@{}}
\toprule
Model & Correctness (\%) & Groundedness (\%) & Retrieval accuracy (\%) \\
\midrule

\multicolumn{4}{@{}l}{\textbf{Embedders}} \\
\addlinespace[2pt]
\hspace{0.8em}Kanon 2 Embedder       & \textbf{94.0} & \textbf{96.0} & \textbf{86.0} \\
\hspace{0.8em}Text Embedding 3 Large & 76.5          & 91.5          & 52.0          \\
\hspace{0.8em}Gemini Embedding 001   & 74.0          & 87.0          & 53.0          \\
\midrule

\multicolumn{4}{@{}l}{\textbf{LLMs}} \\
\addlinespace[2pt]
\hspace{0.8em}Gemini 3.1 Pro         & \textbf{82.3} & \textbf{94.3} & \textbf{63.7} \\
\hspace{0.8em}GPT-5.2                & 80.7          & 88.7          & \textbf{63.7} \\
\bottomrule
\end{tabularx}
\captionsetup{width=\TableWidth, skip=6pt, justification=centering}
\caption{Average embedding model and LLM dimension scores.}
\label{tbl:avg-emb-llm-dims}
\end{table}

From these results, it is clear that choice of embedding model, not LLM, dominates the performance of legal RAG systems. Kanon 2 Embedder, the best-performing embedding model, in particular, achieves 11.7 points higher correctness, 1.7 points higher groundedness, and 22.3 points higher retrieval accuracy than Gemini 3.1 Pro, the best-performing LLM. Switching to OpenAI's Text Embedding 3 Large correspondingly drops correctness by 17.5 points, groundedness by 4.5 points, and retrieval accuracy by 34 points.

We further observe that choice of LLM has a moderate effect on correctness and groundedness. Gemini 3.1 Pro on average achieves 1.6 points higher correctness than GPT-5.2 and 5.6 points higher groundedness. Notably, when Kanon 2 Embedder is deployed alongside Gemini 3.1 Pro and GPT-5.2, the gap in groundedness closes significantly—GPT-5.2 ends up coming 2 points ahead of Gemini 3.1 Pro. This suggests that GPT-5.2 may be much more likely to hallucinate when provided with irrelevant context.

\subsection{Error decomposition}
\begin{figure}[!htbp]
  \centering
  \includegraphics[width=\linewidth]{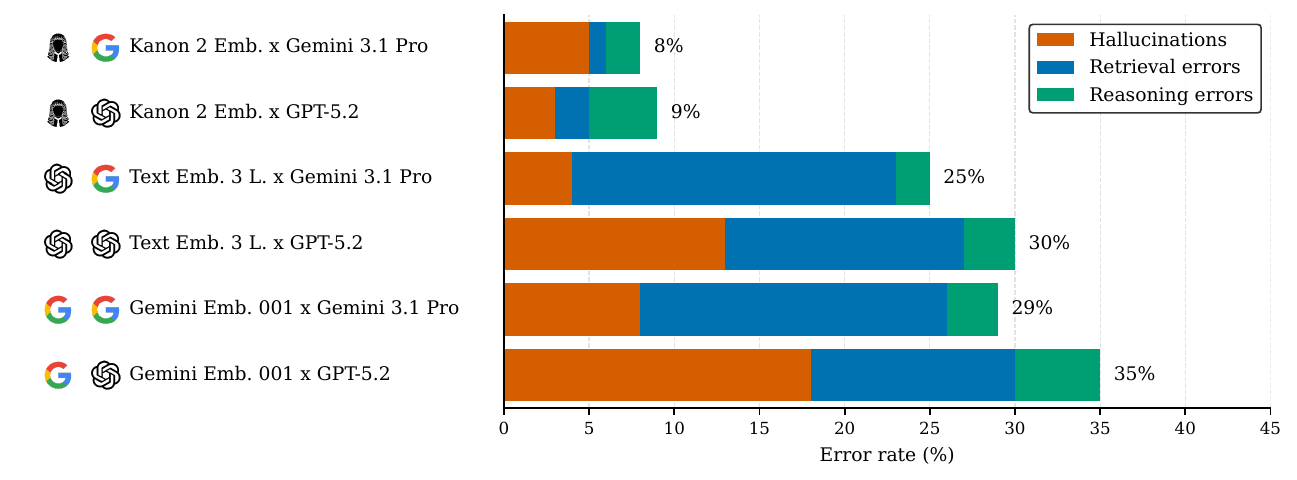}
  \caption{Decomposed error rates for each combination of embedding model and LLM.}
  \label{fig:decomp-err-rates}
\end{figure}

Leveraging our novel error decomposition framework, we go beyond merely reporting correctness, groundedness, and retrieval accuracy and further decompose errors into hallucinations, reasoning failures, and retrieval failures as reported in \autoref{fig:decomp-err-rates}. This decomposition allows us to triangulate the precise origin of errors. 

\FloatBarrier

One of the most important findings of our evaluation is that poor retrieval tends to correlate strongly with an increase in hallucinations. Relative to Text Embedding 3 Large, Gemini Embedding 001 results in a 4.5-point increase in hallucinations on average. Kanon 2 Embedder, in turn, results in a 6.75-point average decrease in hallucinations compared to its general-purpose alternatives. We infer from this that generative models may be able to tell when a passage is more likely to be correct or at least relevant and, in such circumstances, are less likely to invent new facts to help them provide an answer.

Switching generative models also has a moderate effect on hallucinations, with Gemini 3.1 Pro having an average hallucination rate of 5.7\% and GPT-5.2 having a rate of 11.3\%.

Generative errors are proportionally higher with Kanon 2 Embedder; however, that is likely because a dramatic reduction in upstream retrieval errors shifts failures into the generative layer of RAG pipelines. In other words, instead of poor retrieval models handicapping the end RAG performance of generative models, it is now inferior generative models that are handicapping the end RAG performance of high-quality retrieval models.

\begin{figure}[!htbp]
  \centering
  \includegraphics[width=\linewidth]{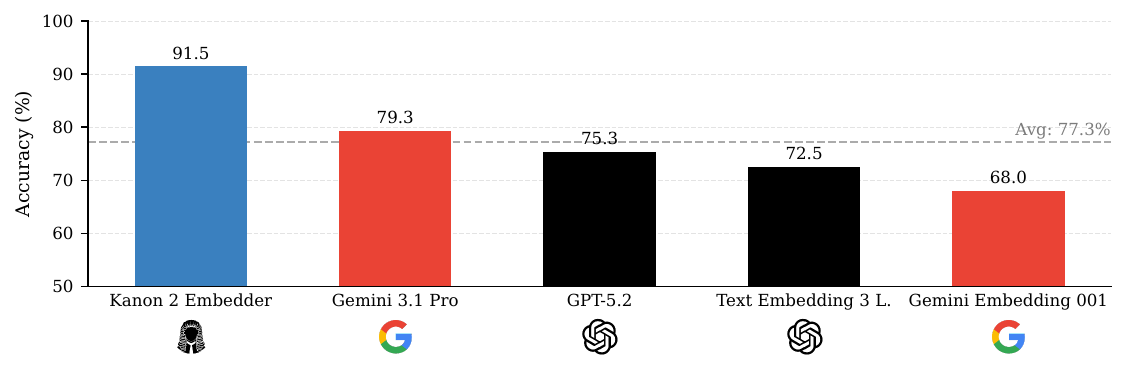}
  \caption{Model impact on RAG accuracy relative to sample average.}
  \label{fig:rag-accuracy}
\end{figure}

We also measure each model’s average RAG accuracy as reported in \autoref{fig:rag-accuracy}, enabling apples-to-apples comparisons between embedding and generative models. We define RAG accuracy to mean the complement of the sum of all errors. To our knowledge, this analysis constitutes the most robust comparison of the effects of embedding and generative models on end legal RAG performance yet.

\FloatBarrier

Our results show that Kanon 2 Embedder delivers 18\% better overall RAG accuracy relative to the sample average. GPT-5.2 and Gemini 3.1 Pro, by contrast, impact accuracy by -3\% and +3\%, respectively. These results reinforce that choice of embedding model is the primary driver behind RAG performance, revealing an even more pronounced effect when false-positive hallucinations are accounted for.

\subsection{Statistical significance and interaction effects}
An experiment may reveal a positive relationship without establishing that relationship as causal or statistically significant. In factorial experiments like ours, there is an additional concern: differences in scores may be driven by interaction effects rather than the stand-alone performance of each model.

One plausible channel for such interactions lies in the selection of embedding and generative models included in our sample. For example, Gemini Embedding 001 may be unusually well matched with Gemini 3.1 Pro, since both are products of the same model builder, Google. If so, we cannot claim that Gemini 3.1 Pro is generally better at Legal RAG Bench than GPT-5.2 simply because it performs better in this experiment. In a different experiment—for example, one that replaces Gemini Embedding 001 with Cohere Embed v4—GPT-5.2 might outperform Gemini 3.1 Pro instead.

Despite the threat interaction effects pose to the external validity of legal evaluations, most evaluators neither test for nor report interaction effects. Uniquely, we do exactly that.

To quantify main effects, test for interactions, and account for estimate uncertainty, we fit a linear probability model for $y\in\{c_{eli}, g_{eli}, b_{eli}\}$, where $b_{eli}=1$ if $c_{eli} \wedge g_{eli}$:
\[
y_{eli} = \alpha_i + \beta_e + \gamma_l + \delta_{el} + \varepsilon_{eli}.
\]
Here, $\alpha_i$ denotes question fixed effects, $\beta_e$ and $\gamma_l$ are embedding model and LLM indicators, and $\delta_{el}$ captures embedder--LLM interactions. We estimate our model using ordinary least squares and report cluster-robust standard errors clustered by question.

We further report ANOVA-style Wald tests aligned with our factorial design:
\begin{itemize}
  \item \textbf{Interaction effects:} test $H_0: \delta_{el}=0$ for all embedder--LLM interaction terms. Under our baselines, this is a joint test that the GPT-5.2 interaction coefficients for the non-baseline embedding models (Gemini Embedding 001 and Kanon 2 Embedder) are zero.
  \item \textbf{LLM main effect.} Test the average simple effect of switching from Gemini 3.1 Pro to GPT-5.2 across all embedding models.
  \item \textbf{Embedding model main effects.} Test embedding model contrasts relative to the baseline embedding model (Text Embedding 3 Large) averaged across LLMs. Concretely, for each non-baseline embedding model $e$, we test whether $\beta_e$ differs from zero under this averaging scheme.
\end{itemize}

\autoref{tbl:chi2_results} reports ANOVA-style Wald tests from the linear probability model with question fixed effects and cluster-robust standard errors clustered by question. Across all three outcomes, embedding model main effects are statistically significant, indicating that choice of embedding model accounts for a substantial share of performance variation. By contrast, the LLM main effect is not statistically distinguishable from zero for correctness or for both correctness and groundedness.

\FloatBarrier

\begin{table}[htb]
\centering

\footnotesize
\setlength{\tabcolsep}{6pt}
\setlength{\extrarowheight}{2pt}
\renewcommand{\arraystretch}{1.30}

\begin{tabularx}{\linewidth}{@{}>{\RaggedRight\arraybackslash}p{0.28\linewidth} *{3}{>{\centering\arraybackslash}X}@{}}
\toprule
 & \textbf{Correct} & \textbf{Grounded} & \textbf{Both} \\
\midrule
\rowlabel{Interaction} &
\statcellNS{$\chi^2(2)=0.08$}{$(p=0.962)$} &
\statcellSig{$\chi^2(2)=8.12$}{$(p=0.017^{*})$} &
\statcellNS{$\chi^2(2)=1.47$}{$(p=0.479)$} \\
\rowlabel{LLM main effect} &
\statcellNS{$\chi^2(1)=0.46$}{$(p=0.499)$} &
\statcellSig{$\chi^2(1)=5.36$}{$(p=0.021^{*})$} &
\statcellNS{$\chi^2(1)=2.43$}{$(p=0.119)$} \\
\rowlabel{Embedder main effect} &
\statcellSig{$\chi^2(2)=28.63$}{$(p<0.001^{***})$} &
\statcellSig{$\chi^2(2)=8.95$}{$(p=0.011^{*})$} &
\statcellSig{$\chi^2(2)=32.00$}{$(p<0.001^{***})$} \\
\bottomrule
\end{tabularx}
\emph{Note}: Stars: $^{***}p<0.001$, $^{**}p<0.01$, $^{*}p<0.05$, $^{\dagger}p<0.1$. Bold indicates $p<0.05$.
\captionsetup{skip=8pt}
\caption{ANOVA-style Wald $\chi^2$ tests from the linear probability model $y_{eli}=\alpha_i+\beta_e+\gamma_l+\delta_{el}+\varepsilon_{eli}$ with question fixed effects and cluster-robust SEs clustered by question, reported separately for $c_{eli}$ (correct), $g_{eli}$ (grounded), and $b_{eli}$ (both; $b_{eli}=1$ iff $c_{eli}=g_{eli}=1$).}
\label{tbl:chi2_results}
\end{table}

The only outcome for which choice of LLM matters on average is groundedness, where the LLM main effect is significant. Interestingly, \autoref{tbl:chi2_results} also shows that groundedness is the only metric with detectable embedding model--LLM interactions. This implies that differences in groundedness are not stable across embedding models, and should not be summarized by a single global LLM ranking. More generally, these results motivate reporting interaction tests alongside main effects when benchmarking RAG pipelines.

\begin{table}[htbp]
\centering
\small
\setlength{\tabcolsep}{6pt}
\renewcommand{\arraystretch}{1.15}
\begin{tabular}{llcc}
\toprule
\textbf{Effect} & \textbf{Contrast} & \textbf{Est. (pp) (SE)} & \textbf{$\bm{p}$} \\
\midrule
$\Delta$LLM & with Text Emb. 3 L.   & \textbf{-9.0 (4.2)$^{*}$}  & \textbf{0.031} \\
$\Delta$LLM & with Gem. Emb. 001  & \textbf{-10.0 (4.8)$^{*}$} & \textbf{0.038} \\
$\Delta$LLM & with Kanon 2 Emb.      & 2.0 (2.7)                & 0.459 \\
\addlinespace
$\Delta\Delta$ & Kanon 2 Emb. $-$ Text Emb. 3 L.    & \textbf{11.0 (5.2)$^{*}$} & \textbf{0.033} \\
$\Delta\Delta$ & Kanon 2 Emb. $-$ Gem. Emb. 001   & \textbf{12.0 (5.0)$^{*}$} & \textbf{0.017} \\
$\Delta\Delta$ & Gem. Emb. 001 $-$ Text Emb. 3 L. & -1.0 (6.2) & 0.871 \\
\bottomrule
\end{tabular}

\vspace{4pt}
\footnotesize
\emph{Note}: $\Delta$LLM is the simple effect of switching the generator from Gemini 3.1 Pro to GPT-5.2 within an embedder.
$\Delta\Delta$ is the corresponding difference-in-differences relative to the baseline embedder (Text Embedding 3 L.).
Stars: $^{***}p<0.001$, $^{**}p<0.01$, $^{*}p<0.05$, $^{\dagger}p<0.1$. Bold indicates $p<0.05$.
\caption{Simple LLM effects and embedder--LLM interaction contrasts for groundedness ($g_{eli}$).}\label{tbl:groundedness_contrasts}
\end{table}

\autoref{tbl:groundedness_contrasts} unpacks the groundedness interaction further by reporting the simple effect of switching generative models for each embedding model, along with interaction contrasts relative to the baselines. Switching from Gemini 3.1 Pro to GPT-5.2 reduces groundedness by 9.0 points when used with Text Embedding 3 Large and by 10.0 points with Gemini Embedding 001, but produces no statistically detectable change with Kanon 2 Embedder. The interaction contrasts indicate that this heterogeneity is driven by Kanon 2 Embedder. The LLM switch effect with Kanon 2 Embedder is 11 points more positive than with Text Embedding 3 Large and 12 points more positive than with Gemini Embedding 001, while Text Embedding 3 Large and Gemini Embedding 001 do not differ meaningfully.

Taken together, \autoref{tbl:chi2_results} and \autoref{tbl:groundedness_contrasts} reinforce our main conclusions. Choice of embedding model is the dominant and most statistically robust driver of end-to-end legal RAG performance. Hallucinations are strongly shaped by retrieval, although this relationship can be obscured by embedding model–LLM interactions. Finally, by testing interactions alongside main effects, we obtain component-level estimates and show that, for correctness, embedding model and LLM performance gains are linearly additive. Practically, this means that selecting the best-performing embedding model and LLM on Legal RAG Bench yields a predictable improvement when the components are deployed together over a distributionally similar corpus.

\section{Conclusion}
Legal RAG systems are deployed in settings where groundedness matters just as much, if not more, than correctness. Verifiably incorrect conclusions can always be proved to be false, whereas unverifiably correct conclusions can never be proven to be true. As RAG becomes increasingly indispensable to modern legal AI applications, it is therefore important that practitioners rely on legal benchmarks that are not only high quality and correlated with real-world performance but also evaluate the failure modes that matter most, including hallucinations and reasoning failures. Until now, practitioners have been let down by benchmarks that are often lacking in quality, methodological soundness, and relevance to real legal work. We address this scarcity in reliable legal RAG evaluations by introducing Legal RAG Bench.

As a dataset, Legal RAG Bench pairs 4,876 passages from the Victorian Criminal Charge Book \cite{jcv_criminal_charge_book} with 100 meaningfully challenging expert-crafted questions and long-form answers. As a methodology, Legal RAG Bench assesses RAG pipelines end-to-end, leveraging a full factorial design and novel hierarchical error decomposition taxonomy. In effect, Legal RAG Bench constitutes one of the most useful and robust evaluations of the real-world performance of legal RAG systems.

Our results conclusively establish that across all evaluation dimensions, choice of embedding model dominates RAG performance. This finding is made concrete through a comprehensive analysis of its statistical significance and potential interaction effects. We further establish that poor information retrieval can often trigger an increase in legal hallucinations, suggesting that generative models may potentially be aware of when they are hallucinating invented facts to answer questions to which they have no relevant facts. Switching to a high-quality domain-adapted legal embedding model such as Kanon 2 Embedder can, therefore, raise the ceiling on performance to the point where the quality of reasoning models bottlenecks pipelines rather than embedding models.

In the interests of transparency, we openly release the code, data, and results behind Legal RAG Bench. We have also added Legal RAG Bench to MLEB\cite{butler2025mleb}, further improving the state of legal information retrieval evaluations more generally. We encourage reproductions of our results.

\section{Data Availability}
Legal RAG Bench is publicly available as a dataset on Hugging Face at \url{https://huggingface.co/datasets/isaacus/legal-rag-bench}. The code behind Legal RAG Bench is available on GitHub at \url{https://github.com/isaacus-dev/legal-rag-bench}. Finally, we present all of the results of Legal RAG Bench in an interactive viewer in our blog post announcing it: \url{https://isaacus.com/blog/legal-rag-bench}.

\section{Disclosures}
Isaacus, a foundational legal AI research company, of which the authors are the founders, created Kanon 2 Embedder and sponsored the creation of Legal RAG Bench and MLEB \cite{butler2025mleb}.

\bibliographystyle{plain}
\bibliography{sources}
\end{document}